%% file: main.tex
\documentclass[11pt,onecolumn]{article}
\usepackage{spconf,amsmath,epsfig}
\usepackage{hyperref}
\usepackage{amsmath}
\usepackage{cite}
\usepackage{amssymb}
\usepackage{algorithm}
\usepackage{algpseudocode}
\usepackage{balance}
\usepackage{hyperref}
\hypersetup{
    bookmarks=true,         % show bookmarks bar?
    unicode=false,          % non-Latin characters in Acrobats bookmarks
    pdftoolbar=true,        % show Acrobats toolbar?
    pdfmenubar=true,        % show Acrobats menu?
    pdffitwindow=false,     % window fit to page when opened
    pdfstartview={FitH},    % fits the width of the page to the window
    pdftitle={sparse DTM},    % title
    pdfauthor={Camps-Valls},     % author
    pdfsubject={sparse DTM},   % subject of the document
    pdfcreator={Camps-Valls},   % creator of the document
    pdfproducer={Camps-Valls}, % producer of the document
    pdfkeywords={keyword1} {key2} {key3}, % list of keywords
    pdfnewwindow=true,      % links in new window
    colorlinks=true,       % false: boxed links; true: colored links
    linkcolor=blue,          % color of internal links (change box color with linkbordercolor)
    citecolor=blue,        % color of links to bibliography
    filecolor=blue,      % color of file links
    urlcolor=blue           % color of external links
}
\newcommand{\matr}[1]{\mathbf{#1}}

\title{Sparsity-driven Digital Terrain Model Extraction}
\input{authors}

\begin{document}

\maketitle

\input{abstract}
\input{keywords}
\input{introduction}

\input{method}

\input{experiments}

\input{conclusion}

\end{document}

%% file: authors.tex
\name{Fatih Nar$^{1}$, 
  	  Erdal Yilmaz$^{2}$, 
      Gustau Camps-Valls$^{3}$
\thanks{{\bf Preprint. Paper published in IGARSS 2018 - 2018 IEEE International Geoscience and Remote Sensing Symposium, Valencia, 2018, pp. 1316-1319, doi: 10.1109/IGARSS.2018.8517569.}}
	\thanks{This work was supported by the Scientific and Technical Research Council of Turkey (TUBITAK), Grant Number: TUBITAK-BIDEB-2219. GCV was funded by the European Research Council (ERC) under the ERC-CoG-2014 SEDAL project (grant agreement 647423). Authors would like to thank A. Ozgur and M. Ergul for their help during the preliminary investigation.}
}
\address{$^{1}$Konya Food and Agriculture University, Konya, Turkey; $^{2}$Zibumi Studios, Ankara, Turkey \\
$^{3}$Image Processing Lab (IPL), Universitat de Val\`encia, Val\`encia, Spain}

%% file: abstract.tex
\begin{abstract}
We here introduce an automatic Digital Terrain Model (DTM) extraction method. 
The proposed sparsity-driven DTM extractor (SD-DTM) takes a high-resolution Digital Surface Model (DSM) as an input and constructs a high-resolution DTM using the variational framework. 
To obtain an accurate DTM, an iterative approach is proposed for the minimization of the target variational cost function.  
Accuracy of the SD-DTM is shown in a real-world DSM data set. 
We show the efficiency and effectiveness of the approach both visually and quantitatively via residual plots in illustrative terrain types.
\end{abstract}

%% file: keywords.tex

%% file: introduction.tex
\section{Introduction}
\label{section:introduction}
\begin{figure}[h]
\vspace*{-1 mm}
\centering{\includegraphics[width=0.95\columnwidth]{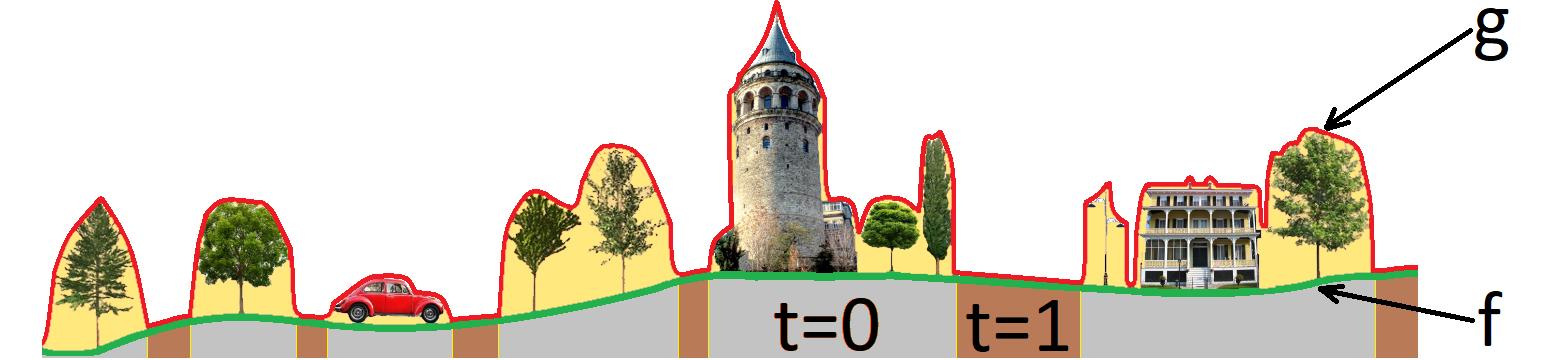}}
\vspace{0pt}
\vspace*{-3 mm}
\caption{DSM versus DTM.}
\label{DSMversusDTM}
\end{figure}

A Digital Terrain Model (DTM) is an elevation map of bare ground where man-made objects (buildings, vehicles, etc.) as well as vegetation  (trees, bushes, etc.) are removed from the Digital Surface Model (DSM)~\cite{li2004digital}. 
In Fig.\ref{DSMversusDTM}, $g$ represents surface elevations hence DSM, $f$ represents terrain elevations hence DTM, and $t$ represents terrain vs non-terrain classification ($t=1$ for terrain regions, $t=0$ for non-terrain regions). 

DTMs are useful for extracting man-made and vegetation objects, extracting terrain parameters, precision farming and forestry, planning of new roads and railroads, visualization and simulation of the 3D world, modeling physical phenomenas, such as water flow or mass movement, rectification of aerial photography or satellite imagery, and many other Geographic Information Systems (GIS) tasks~\cite{Lee2003,Zhang2003,li2004digital, Unger2009,Beumier2016}. 
However, manual preparation of a DTM using ground measurements is expensive and time consuming~\cite{Lee2003}. 
Certainly, the definition of DTM is often elusive and controversial. 
Thus, automatic extraction of a DTM from an automatically obtainable DSM is a reasonable and often preferred alternative, even though it poses important challenges to be addressed~\cite{George2004,Joachim2009}. 

Several approaches to derive DTM exist in the literature. 
In~\cite{Lee2003}, a modified linear prediction technique followed by adaptive processing is proposed for DTM extraction.
In~\cite{Zhang2003}, a progressive morphological filter is developed to preserve ground while removing non-ground objects. 
An alternative approach was presented in~\cite{Unger2009}, where a variational approach is proposed for the semiautomatic generation of the DTM. 
More recently, in \cite{Domen2014}, the most contrasted connected-components are extracted to generate DTM from LiDAR data, while in~\cite{Beumier2016}, the DSM is segmented into uniform regions and interpolation is applied between selected regions. 
Lately, in \cite{Abdullah2017}, 2D empirical mode decomposition is proposed for DTM generation. 

In this work we propose, a methodology based on the variational approach introduced in~\cite{Unger2009}. 
Our proposed methods follows an iterative procedure that `peels the onion' according to a target cost function under sparsity-preserving constraints. 
Accuracy of the derived DTM will be shown in a real-world DSM data set, and analyzed both qualitatively and quantitatively in illustrative terrain types.
 
The remainder of the paper is organized as follows.
section \ref{section:proposed_method} briefly reviews the proposed method used in this work. 
Section \ref{section:experimental_results} first describes the dataset collected, and then gives an empirical evidence of performance both visually and quantitatively. 
We conclude in section \ref{section:conclusion} with some remarks and an outline future work.

%% file: method.tex
\section{Proposed DTM Extraction Method}
\label{section:proposed_method}

DTM can be constructed from a DSM by interpolating the elevation values in the non-terrain cells using the elevations of the nearby terrain cells~\cite{Beumier2016}.
However, manual delineation of the cells (as terrain versus non-terrain) is a tedious task~\cite{Lee2003} error-prone, and automatic classification is challenging \cite{Camps-Valls2007,Camps-Valls2012}. 
On top of all this, determining the elevation values for the non-terrain cells is an ill-posed scattered data interpolation problem, where it is also sensitive to errors in the terrain non-terrain boundaries \cite{li2004digital}.

Inspired by \cite{Unger2009}, to handle the above mentioned issues, we propose the minimization of a similar variational cost function, yet by using a novel iterative approach and numerical solver for the construction of DTM. 
The pseudocode is given in Algorithm \ref{DTM_Extraction_Pseudocode}: firstly the DTM ($f$) is initialized with elevations of the DSM ($g$), then a terrain indicator map ($t$) is updated which is followed by an update of the terrain elevation values in an iterative manner. 
The algorithm is iterated with the previous solution until it convergences or a maximum number of iterations $n_{max}$ is reached. 
In this study, we use a regular grid format for representing the DSM and the DTM, where each grid cell stores a floating number for its elevation value.

\begin{algorithm}
\caption{DTM Extraction Pseudo-code}
\label{DTM_Extraction_Pseudocode}
\begin{algorithmic}[1]
  \State \textbf{Input:} $g, n_{max}$
  \State \textbf{Initialize:} $f^{(1)} \gets g$
  
  \For{$n=1$ to $n_{max}$}
		\State{Update terrain indicator map $t^{(n)}$ using $f^{(n)}$ and $g$}
		\State{Update terrain elevations $f^{(n+1)}$ using $t^{(n)}$, $f^{(n)}$, $g$}
        \State{Check for convergence using $f^{(n)}$ and $f^{(n+1)}$}
  \EndFor
  
  \State \textbf{return} $f$
\end{algorithmic}
\end{algorithm}

\begin{figure}[h]
\centering{\includegraphics[width=0.99\columnwidth]{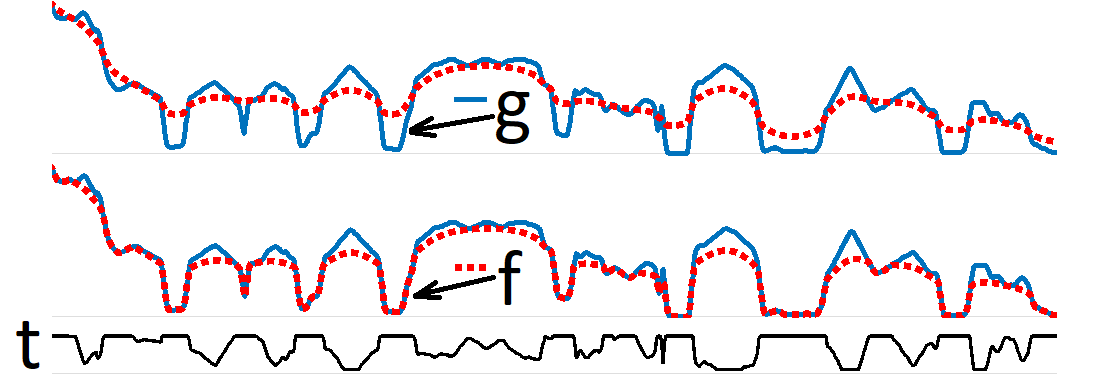}}
\vspace*{-1 mm}
\caption{DSM versus DTM.}
\label{constraiedVersusUnconstrained}
\end{figure}

If DSM is smoothed, then elevations of non-terrain objects will become lower. 
However, this simple approach also leads to an increase in the elevations for the terrain regions (see Fig.~\ref{constraiedVersusUnconstrained}, top). 
In order to prevent this problem, smoothing can be applied onto the DSM using the prior knowledge $f \leqslant g$. 
This prior knowledge can be included in the minimization functional as an inequality constraint, and thus can be combined into a smoothing operation by the minimization of a cost function which will prevent the height increase in terrain regions. (see Fig.~\ref{constraiedVersusUnconstrained}, middle).
In Fig.~\ref{constraiedVersusUnconstrained}, solid blue line is surface ($g$) where dotted red line is smoothed surface ($f$).

If we define $f$ as the smoothed version of the surface $g$, then the terrain indicator map for each cell can be defined as below (see Fig.~\ref{constraiedVersusUnconstrained}, bottom):
\begin{equation}
\label{TerrainIndicatorFunction}
\begin{aligned}
	t_p = 1 - \min\bigg(\frac{1}{T_{ng}}(g_p - f_p), 1\bigg),
\end{aligned}
\end{equation}
where $p$ is the cell index number, $t$ is the terrain indicator map, $g$ is the existing DSM, $f$ is the smoothed DSM (rough DTM), $T_{ng}$ is a terrain threshold (set to $0.5$ for simplicity).

In this study, the proposed variational cost function that is minimized to obtain terrain elevations ($f$) by smoothing the surface elevations ($g$) using the prior ($f \leqslant g$) and the terrain indicator map ($t$) as following:
\begin{equation}
\label{InitialCostFunction}
\begin{aligned}
	J(f) =  \sum_{p=1}^{} t_p((|f_p - g_p|+1)^2-1) + \lambda|(\nabla{f})_p| 
    \\ \textrm{w.r.t.}  ~~ f_p \leqslant g_p,
\end{aligned}
\end{equation}
where $p$ is the cell index number, $t$ is the terrain indicator map, $g$ is the existing DSM, $f$ is the DTM to be obtained, $\lambda$ is a positive value determining smoothing level, and $\nabla$ is the gradient operator. The first term is the {\em data fidelity term} that ensures keeping $f$ similar to $g$ by using an $\ell_1$-norm penalty when the difference between $f$ and $g$ is small, and an $\ell_2$-norm as the difference gets larger. 
The second term is the total variation (TV) {\em regularization term} that implies a penalty on the changes in image gradients using an $\ell_1$-norm, thus preserving details while  enforcing smoothness~\cite{rudin1992nonlinear}.
Higher smoothing effects are obtained by an increasing the $\lambda$ value. 
The constraint, $f_p \leqslant g_p$, prevents terrain elevations being higher than surface elevations, as common sense dictates.
Here, $t$ indicates a fuzzy membership ($0 \leqslant t_p \leqslant 1$) such that $t_p=0$ for non-terrain cells and $t_p=1$ for terrain cells. 
As $t_p$ gets closer to $0$, data fidelity term vanishes and only TV-regularization (TV diffusion) term remains, thus cost function acts as a scattered data interpolator.
As $t_p$ gets closer to $1$, data fidelity term becomes active and surface is preserved more.

\subsection{Minimization of the cost function}
\label{section:minimization_of_the_cost_function}

After doing algebraic manipulations and taking the constraint, $f_p \leqslant g_p$, into the cost function using the penalty method with $\lambda_p$ as penalty multiplier, equation (\ref{InitialCostFunction}) becomes as below:
\begin{equation}
\label{ConstraintAsPenaltyCostFunction}
\begin{aligned}
	J(f) =  \sum_{p=1}^{} & t_p((f_p - g_p)^2+2|f_p - g_p|) \\
    + & \lambda|(\nabla{f})_p| + \lambda_p max(f_p - g_p, 0)
\end{aligned}
\end{equation}
In equation (\ref{ConstraintAsPenaltyCostFunction}), maximum function ($max$) returns zero penalty if $f_p \leqslant g_p$ and it returns a penalty proportional to $\lambda_p$ otherwise.
$\lambda_p$ should be increased as the smoothing ($\lambda$) increases, thus we set $\lambda_p = 0.5\lambda$. 

Although equation (\ref{ConstraintAsPenaltyCostFunction}) is convex, absolute and $max$ functions are non-differentiable which makes the minimization difficult.
Inspired from \cite{Ozcan2016,Nar2016}, we set $\hat{f}_p$ as a proxy for $f_p$ to be able to approximate non-differentiable terms in equations \eqref{AbsoluteFidelityApproximation}, \eqref{TVRegularizationApproximation}, and \eqref{MaxApproximation}. 
First, the absolute function in the data fidelity term is approximated as below:
\begin{equation}
\label{AbsoluteFidelityApproximation}
\begin{aligned}
	|f_p - g_p| \approx & ~ d_p(f_p - g_p)^2 \\
    					& ~ d_p = (|\hat{f}_p - g_p| + \varepsilon)^{-1},
\end{aligned}
\end{equation}
where $\varepsilon$ is a small positive constant. In this study, $\varepsilon=0.1$ is used for all the experiments. Second, the absolute value of the gradient operator is approximated as:
\begin{equation}
\label{TVRegularizationApproximation}
\begin{aligned}
	|(\nabla{f})_p| & = |(\partial_x{f})_p| + |(\partial_y{f})_p| \\
     				& \approx w_{x,p} {(\partial_x{f})_p^2}  + w_{y,p} {(\partial_y{f})_p^2} \\
          & ~~~ w_{x,p} = (|(\partial_x{\hat{f}})_p| + \varepsilon)^{-1} \\ 
          & ~~~ w_{y,p} = (|(\partial_y{\hat{f}})_p| + \varepsilon)^{-1}
\end{aligned}
\end{equation}
Finally, the max-function is approximated as:
\begin{equation}
\label{MaxApproximation}
\begin{aligned}
	\max & (f_p - g_p, 0) \approx h_p(f_p - g_p)^2 \\
   	& h_p = sgn(\max(\hat{f}_p - g_p, 0)) d_p
\end{aligned}
\end{equation}
where $sgn$ is the sign function.
The approximated cost function in equation \eqref{ApproximatedCostFunction} is accurate around $\hat{f}_p$ so it must be solved in an iterative manner \cite{Nar2016}, where $n$ is the iteration number. This cost function has a different data fidelity term and numerical minimization approach and it is also iterative comparing to two-phase solution proposed in \cite{Unger2009}.
\begin{equation}
\label{ApproximatedCostFunction}
\begin{aligned}
	J^{(n)}(f) =  \sum_{p=1}^{} & t_p((f_p - g_p)^2 + 2d_p(f_p - g_p)^2) \\
    + & \lambda(w_{x,p} {(\partial_x{f})_p^2} + w_{y,p} {(\partial_y{f})_p^2}) \\
    + & \lambda_p h_p(f_p - g_p)^2
\end{aligned}
\end{equation}
Equation \eqref{ApproximatedCostFunction} can be cast in the matrix-vector form as below:
\begin{equation}
\label{ApproximatedCostFunctionMatrixVectorForm}
\begin{aligned}
	J^{(n)}(v_f) = & \Big( (v_f - v_g)^\top + 2(v_f - v_g)^\top \matr{D} \Big) \matr{T} (v_f - v_g) \\
    			+ & \lambda(v_f^\top \matr{C_x}^\top \matr{W_x} \matr{C_x} v_f + v_f^\top \matr{C_y}^\top \matr{W_y} \matr{C_y} v_f) \\
                + & \lambda_p(v_f - v_g)^\top \matr{H} (v_f - v_g),
\end{aligned}
\end{equation}
where $v_g$, $v_f$, and $v_{\hat{f}}$ are vector forms of $g_p$, $f_p$, and $\hat{f}_p$; $\matr{D}$ is a diagonal matrix formed of $d_p$; $\matr{T}$ is a diagonal matrix with entries $t_p$, $\matr{H}$ is a diagonal matrix formed of $h_p$; $\matr{W_x}$, $\matr{W_y}$ are diagonal matrices with entries $w_{x,p}$, $w_{y,p}$; 
and $\matr{C_x}$, $\matr{C_y}$ are the Toeplitz matrices as the forward difference gradient operators with zero derivatives at the right and bottom boundaries.

Equation \eqref{ApproximatedCostFunctionMatrixVectorForm} is quadratic; and hence taking its derivatives with respect to $v_f$ and equating to zero yields its global minimum. This leads to the below sparse linear system:
\begin{equation}
\begin{aligned}
	\label{DTM_extraction_linear_system}
	\matr{A} &v_f^{(n+1)} = b \\
    \matr{A} &= \matr{R} + \lambda (\matr{C_x}^\top \matr{W_x} \matr{C_x} + \matr{C_y}^\top \matr{W_y} \matr{C_y}) \\
b &= (\matr{R} + \lambda_p \matr{H})v_g,
\end{aligned}			
\end{equation}
where $\matr{R} = \matr{T}(2\matr{D} + \matr{I})$ as $\matr{I}$ being the identity matrix.  
Here, iteration number is $n$ for the $\matr{A}$, $\matr{R}$, $\matr{T}$, $\matr{D}$, $\matr{H}$, $\matr{W_x}$, $\matr{W_y}$ matrices and $b$ vector unless it is explicitly stated.

\subsection{DTM Extraction Algorithm}
\label{section:DTM_extraction_algorithm}

The DTM extraction method is provided in Algorithm \ref{DTM_Extraction_Pseudocode}. Details on the terrain indicator map update and terrain elevations update approaches are given therein.

In Algorithm \ref{DTM_Extraction_Algorithm}, preconditioned conjugate gradient (PCG) with incomplete Cholesky preconditioner (ICP) is used as an iterative solver to solve the linear system at line 9, where the maximum number of PCG iterations is set to $10^3$ and convergence tolerance is set to $10^{-3}$.

\begin{algorithm}
\caption{DTM Extraction Algorithm}
\label{DTM_Extraction_Algorithm}
\begin{algorithmic}[1]
  \State \textbf{Input:} $g, \lambda=5, n_{max}=10^4, C_{tolerance}=10^{-3}$
  \State $v_g \gets g$, $v_f \gets g$, $\lambda_p \gets 0.5\lambda$, $T_{ng} \gets 0.5$, $\varepsilon \gets 0.1$  
  \For{$n=1$ to $n_{max}$}
  		\State \textbf{Update terrain indicator map:}
		\State $v_t = \vec{1} - \min((v_g - v_f) / T_{ng}, \vec{1})$
		\State \textbf{Update terrain elevations:}
		\State $v_{\hat{f}} \gets v_f$
		\State Construct $\matr{W_x}$, $\matr{W_y}$, $\matr{T}$, $\matr{D}$, $\matr{R}$, $\matr{H}$, $\matr{A}$, and $b$
		\State solve $\matr{A} v_f = b$
		\State $v_f \gets \min(v_f, v_g)$  \Comment{force the $f \leqslant g$ constraint}
		\State \textbf{Check for the convergence:}
		\State if $\|v_f - v_{\hat{f}}\|_{\infty} < C_{tolerance}$ then break the loop
  \EndFor
  \State \textbf{return} $f$  where  $f \gets v_f$
\end{algorithmic}
\end{algorithm}

In \cite{Unger2009}, large smoothing factor was used to determine the terrain indicator map, and then the algorithm was executed again with a smaller smoothing factor. 
Alternatively, in our approach, a small smoothing factor is used and the terrain indicator map is iteratively updated, which leads to a better preservation of details in terrain regions. 
Therefore, in Algorithm \ref{DTM_Extraction_Algorithm}, terrain elevations ($f$) are initialized as surface elevations ($g$) and then both the terrain elevations ($f$) and the terrain indicator map ($t$) are iteratively refined.

\begin{figure}[h!]
\centering{\includegraphics[width=0.95\columnwidth]{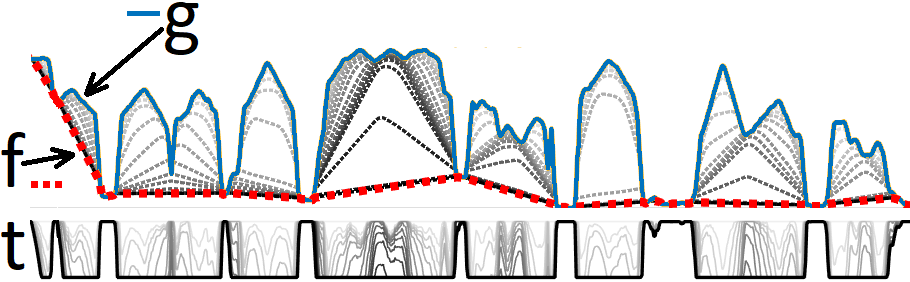}}
\vspace*{-3 mm}
\caption{Evolution of terrain elevations ($f$) and terrain indicator map ($t$) for the Algorithm \ref{DTM_Extraction_Algorithm} on 1-dimensional data.}
\label{dtmEvolution}
\end{figure}
\vspace*{-3 mm}

%% file: experiments.tex
\section{Experimental results}
\label{section:experimental_results}

\subsection{Data Collection and Characteristics}
\label{section:data_collection_characteristics}

We applied the proposed method to Cerkes village dataset to illustrate performance in a large terrain with wide variety of features (i.e. flat regions, hills, rivers, buildings, utility poles, cars, trees, etc.). 
This dataset covers $11~km^2$ area which is obtained using photogrammetry techniques, where raster image has 5 centimeter pixel resolution and DSM has 5 centimeter pixel-spacing. 
Coverage of Cerkes village (at Cankiri city of Turkey) dataset  as a bounding box is given as below:\\ {\footnotesize{N$40^\circ$$49^\prime$$47.07^{\prime\prime}$  E$32^\circ$$52^\prime$$22.17^{\prime\prime}$ ~ to ~ N$40^\circ$$47^\prime$$54.98^{\prime\prime}$  E$32^\circ$$54^\prime$$49.61^{\prime\prime}$}}

\subsection{Visual results}

\begin{figure}[h]
\centering{\includegraphics[width=0.99\columnwidth]{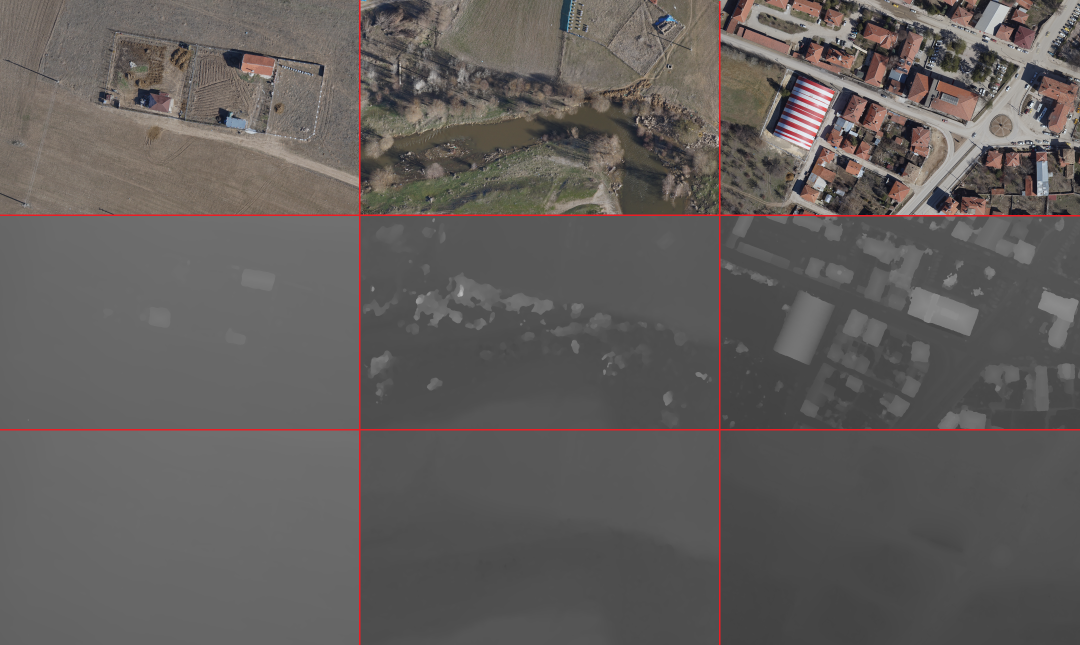}}
\vspace*{-10pt}
\caption{Cerkes data: (a) raster, (b) DSM, (c) extracted DTM.}
\label{RealWorldTestData}
\end{figure}

Fig.~\ref{RealWorldTestData} shows rasters (top), DSMs (middle), and extracted DTMs (bottom) of 3 subregions in the Cerkes village dataset. 
As seen in Fig.~\ref{RealWorldTestData}, man-made objects and vegetations are successfully removed from the terrain and these regions are also interpolated smoothly.
Thus, it can be noted that the proposed method is able to extract bare earth successfully.

\vspace*{-9pt}

\subsection{Numerical evaluation}

\begin{figure}[h]
\centering{\includegraphics[width=0.85\columnwidth]{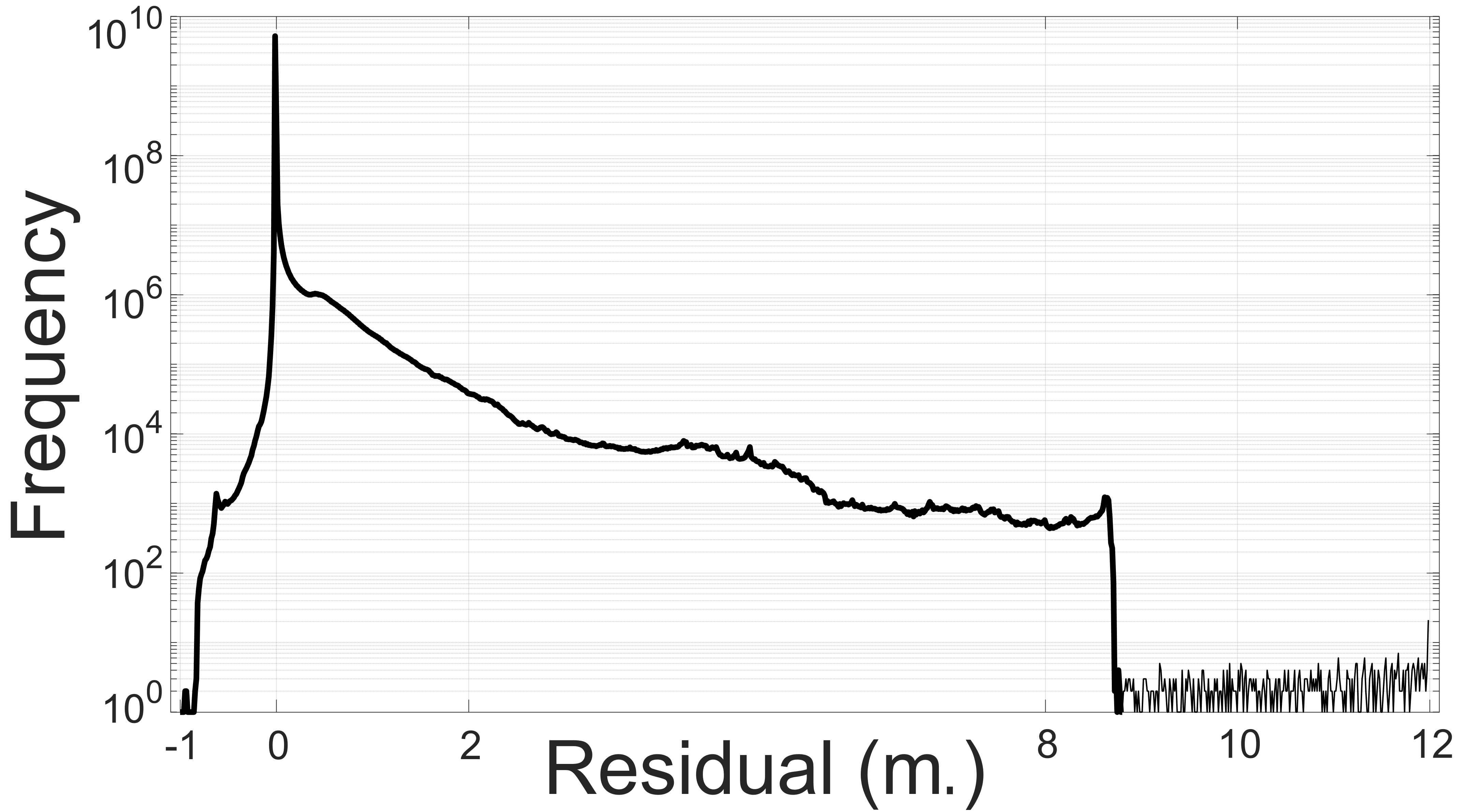}}
\vspace*{-10pt}
\caption{Residual histogram of the proposed method.}
\label{ROCcurve}
\end{figure}

A numerical evaluation was conducted using residual histogram for Cerkes village dataset (in $11~km^2$) where mean residual is $-0.24$ cm, median residual is $0.1$ cm, and mean squared error is $1.19$ cm.
The residual histogram in Fig.~\ref{ROCcurve} shows that the proposed method performs well for a real world data. 
Note that, frequencies of residuals are shown in log$_{10}$-scale to prevent zero residual dominating the plot.

\vspace*{-9pt}

%% file: conclusion.tex
\section{Conclusions}
\label{section:conclusion}

In this study, we presented an automatic DTM extraction method that iteratively estimates terrain indicator map and terrain elevations. 
Experiments show that proposed method can produce an accurate DTM for the given high-resolution DSM where wide variety of non-terrain objects exist on the terrain with various slopes.
Future work will consider adding asymmetry constraints and doing more experiments in regions showing additional characteristics.